\documentclass[10pt,twocolumn,letterpaper]{article}

\usepackage[pagenumbers]{cvpr}

\usepackage{threeparttable}
\usepackage{colortbl}
\usepackage{bbding}
\definecolor{tablegreen}{HTML}{009B55}
\newcommand{\modelicon}[1]{%
    \raisebox{-0.1em}{\includegraphics[width=0.9em,height=0.9em,keepaspectratio]{figures/model_icons/#1.pdf}}\hspace{0.3em}%
}
\newcommand{\PassK}{Pass{\fontencoding{T1}\selectfont\textasciicircum}K}

\definecolor{cvprblue}{rgb}{0.21,0.49,0.74}
\usepackage[pagebackref,breaklinks,colorlinks,allcolors=cvprblue]{hyperref}

\def\confName{CVPR}
\def\confYear{2027}

\makeatletter
\newcommand{\authornote}[1]{\begingroup\let\@makefnmark\@empty\thanks{#1}\endgroup}
\makeatother

\title{MedUIGym: Evaluating and Training GUI Agents for Clinical Workflows}
\author{
\textbf{Yitong Qiao\textsuperscript{1,*,\textdagger}}\authornote{Equal contribution. \href{mailto:qiaoyt@zju.edu.cn}{qiaoyt@zju.edu.cn}; \href{mailto:liulei1497@gmail.com}{liulei1497@gmail.com}.}\authornote{Work done during an internship at Ant Group.},
\textbf{Lei Liu\textsuperscript{1,2,*}},
\textbf{Yue Shen\textsuperscript{2}},
\textbf{Jian Wang\textsuperscript{2}},
\textbf{Jinjie Gu\textsuperscript{2}},
\textbf{Zhixuan Chu\textsuperscript{1,\textdaggerdbl}}\authornote{Corresponding author.}
\\
\\
\textsuperscript{1}The State Key Laboratory of Blockchain and Data Security, Zhejiang University
\\
\textsuperscript{2}Ant Healthcare, Ant Group
\\
\small{
\textbf{Corresponding author:} \href{mailto:zhixuanchu@zju.edu.cn}{zhixuanchu@zju.edu.cn}
}
}

\begin{document}
\addtocontents{toc}{\protect\setcounter{tocdepth}{-1}}
\maketitle
\begin{abstract}

Automating clinical workflows requires vision-language agents that can navigate dense graphical user interfaces (GUIs) in electronic health records (EHRs) and execute precise actions, yet many medical agents still rely on text or structured tool calls. In this work, we introduce \texttt{MedUIGym}, a scalable interactive environment comprising 15,739 tasks from 17 benchmark sources across six clinical task families for training and evaluating clinical GUI agents. In \texttt{MedUIGym}, agents perceive EHR states from screenshots and execute mouse-and-keyboard actions, as well as a paired native-tool interface over identical clinical data. The environment verifies both final answers and required clinical actions, and supports parallel trajectory collection for training. Benchmarking twelve vision-language models shows that current agents struggle across diverse clinical workflows, and simply increasing reasoning effort fails to yield consistent gains.
Furthermore, \texttt{MedUIGym} supports offline learning from GUI trajectories and online reinforcement learning from task rewards.
After training, both Qwen3.5 models perform better on in-domain clinical tasks and generalize to tasks from external sources.
By unifying tasks from existing benchmarks, dual-interface evaluation, and GUI trajectory collection, \texttt{MedUIGym} provides a testbed for studying how vision-language agents complete clinical tasks through the EHR's graphical interface.
Our project page is available at \mbox{\url{https://github.com/yitong-qiao/MedUIGym}}.
\end{abstract}

\section{Introduction}
\label{sec:intro}

\begin{figure}[t]
\centering
\includegraphics[width=\linewidth]{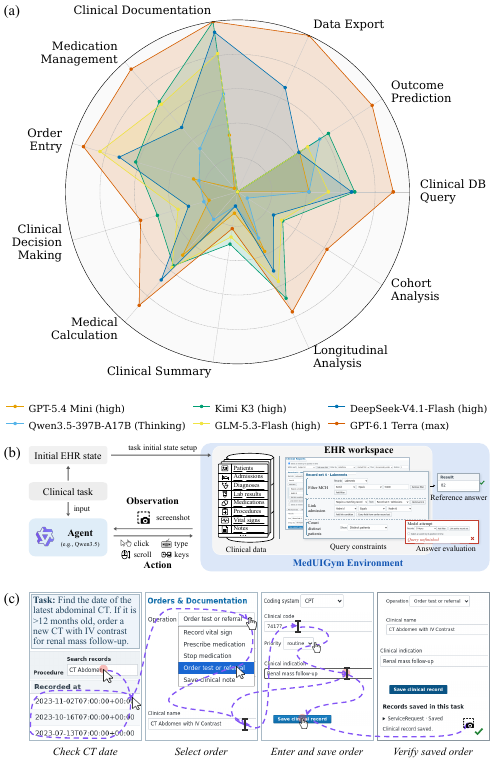}
\caption{\textbf{MedUIGym overview}. (a)~Frontier model performance; (b)~EHR environment and evaluation; (c)~a clinical workflow.}
\label{fig:teaser}
\end{figure}

\begin{table*}[t]
\captionsetup{width=\linewidth,singlelinecheck=false}
\centering
\caption{\textbf{Comparison with related EHR benchmarks}. \texttt{MedUIGym} integrates multi-source clinical tasks, visual GUI interaction, and outcome verification in a unified EHR environment for trajectory sampling and offline and online agent training.}
\label{tab:related}

\begin{threeparttable}
\centering
\renewcommand\arraystretch{1.03}
\setlength{\tabcolsep}{2.5pt}
\resizebox{\linewidth}{!}{%
\begin{tabular}{@{}l|cccccc|cccc|ccc@{}}
\toprule
 & \multicolumn{6}{c|}{\textbf{Task}} & \multicolumn{4}{c|}{\textbf{Environment \& Facility}} & \multicolumn{3}{c}{\textbf{Scale (\#Instances)}} \\
\cmidrule(lr){2-7} \cmidrule(lr){8-11} \cmidrule(lr){12-14}
\textbf{Benchmark} & \shortstack{Information\\Retrieval} & \shortstack{Clinical\\Decision Making} & \shortstack{Cohort\\Queries} & \shortstack{Clinical\\Actions} & \shortstack{Clinical\\Documentation} & \shortstack{Outcome\\Prediction} & \shortstack{Visual\\GUI} & \shortstack{Outcome\\Verification} & \shortstack{Trajectory\\Sampling} & \shortstack{Training\\Resources} & \#Train & \#Test & \#Traj. \\
\midrule
EHRSQL~\citep{lee2022ehrsql} & \textcolor{tablegreen}{\CheckmarkBold} & \textcolor{red}{\XSolidBrush} & \textcolor{tablegreen}{\CheckmarkBold} & \textcolor{red}{\XSolidBrush} & \textcolor{red}{\XSolidBrush} & \textcolor{red}{\XSolidBrush} & \textcolor{red}{\XSolidBrush} & \textcolor{tablegreen}{\CheckmarkBold} & \textcolor{red}{\XSolidBrush} & \textcolor{tablegreen}{\CheckmarkBold} & 18.6K & 6.0K & -- \\
EHR-SeqSQL~\citep{ryu2024ehr} & \textcolor{tablegreen}{\CheckmarkBold} & \textcolor{red}{\XSolidBrush} & \textcolor{tablegreen}{\CheckmarkBold} & \textcolor{red}{\XSolidBrush} & \textcolor{red}{\XSolidBrush} & \textcolor{red}{\XSolidBrush} & \textcolor{red}{\XSolidBrush} & \textcolor{tablegreen}{\CheckmarkBold} & \textcolor{red}{\XSolidBrush} & \textcolor{tablegreen}{\CheckmarkBold} & 6.4K & 2.8K & -- \\
EHR-Complex~\citep{qiao2026ehrcomplex} & \textcolor{tablegreen}{\CheckmarkBold} & \textcolor{red}{\XSolidBrush} & \textcolor{tablegreen}{\CheckmarkBold} & \textcolor{red}{\XSolidBrush} & \textcolor{red}{\XSolidBrush} & \textcolor{red}{\XSolidBrush} & \textcolor{red}{\XSolidBrush} & \textcolor{tablegreen}{\CheckmarkBold} & \textcolor{tablegreen}{\CheckmarkBold} & \textcolor{tablegreen}{\CheckmarkBold} & 48.1K & 3.9K & 6K \\
EHR-RobustGym~\citep{qiao2026ehrrobustgym} & \textcolor{tablegreen}{\CheckmarkBold} & \textcolor{red}{\XSolidBrush} & \textcolor{tablegreen}{\CheckmarkBold} & \textcolor{red}{\XSolidBrush} & \textcolor{red}{\XSolidBrush} & \textcolor{red}{\XSolidBrush} & \textcolor{red}{\XSolidBrush} & \textcolor{tablegreen}{\CheckmarkBold} & \textcolor{tablegreen}{\CheckmarkBold} & \textcolor{tablegreen}{\CheckmarkBold} & 9.0K & 2.0K & 2K \\
MedAgentBench~\citep{jiang2025medagentbench} & \textcolor{tablegreen}{\CheckmarkBold} & \textcolor{tablegreen}{\CheckmarkBold} & \textcolor{red}{\XSolidBrush} & \textcolor{tablegreen}{\CheckmarkBold} & \textcolor{red}{\XSolidBrush} & \textcolor{red}{\XSolidBrush} & \textcolor{red}{\XSolidBrush} & \textcolor{tablegreen}{\CheckmarkBold} & \textcolor{tablegreen}{\CheckmarkBold} & \textcolor{red}{\XSolidBrush} & -- & 300 & -- \\
MedAgentBench v2~\citep{chen2026medagentbenchv2} & \textcolor{tablegreen}{\CheckmarkBold} & \textcolor{tablegreen}{\CheckmarkBold} & \textcolor{red}{\XSolidBrush} & \textcolor{tablegreen}{\CheckmarkBold} & \textcolor{red}{\XSolidBrush} & \textcolor{red}{\XSolidBrush} & \textcolor{red}{\XSolidBrush} & \textcolor{tablegreen}{\CheckmarkBold} & \textcolor{tablegreen}{\CheckmarkBold} & \textcolor{tablegreen}{\CheckmarkBold} & -- & 300 & -- \\
PhysicianBench~\citep{liu2026physicianbench} & \textcolor{tablegreen}{\CheckmarkBold} & \textcolor{tablegreen}{\CheckmarkBold} & \textcolor{red}{\XSolidBrush} & \textcolor{tablegreen}{\CheckmarkBold} & \textcolor{tablegreen}{\CheckmarkBold} & \textcolor{red}{\XSolidBrush} & \textcolor{red}{\XSolidBrush} & \textcolor{tablegreen}{\CheckmarkBold} & \textcolor{tablegreen}{\CheckmarkBold} & \textcolor{red}{\XSolidBrush} & -- & 100 & -- \\
PhysicianUIBench~\citep{anonymous2026physicianuibench} & \textcolor{tablegreen}{\CheckmarkBold} & \textcolor{tablegreen}{\CheckmarkBold} & \textcolor{red}{\XSolidBrush} & \textcolor{tablegreen}{\CheckmarkBold} & \textcolor{tablegreen}{\CheckmarkBold} & \textcolor{red}{\XSolidBrush} & \textcolor{tablegreen}{\CheckmarkBold} & \textcolor{tablegreen}{\CheckmarkBold} & \textcolor{tablegreen}{\CheckmarkBold} & \textcolor{red}{\XSolidBrush} & -- & 100 & -- \\
MedAgentGym~\citep{xu2025medagentgym} & \textcolor{tablegreen}{\CheckmarkBold} & \textcolor{tablegreen}{\CheckmarkBold} & \textcolor{tablegreen}{\CheckmarkBold} & \textcolor{tablegreen}{\CheckmarkBold} & \textcolor{red}{\XSolidBrush} & \textcolor{tablegreen}{\CheckmarkBold} & \textcolor{red}{\XSolidBrush} & \textcolor{tablegreen}{\CheckmarkBold} & \textcolor{tablegreen}{\CheckmarkBold} & \textcolor{tablegreen}{\CheckmarkBold} & 4.3K & 4.8K & 6.7K \\
\midrule
\rowcolor{RoyalPurple!12} \textbf{\texttt{MedUIGym} (Ours)} & \textcolor{tablegreen}{\CheckmarkBold} & \textcolor{tablegreen}{\CheckmarkBold} & \textcolor{tablegreen}{\CheckmarkBold} & \textcolor{tablegreen}{\CheckmarkBold} & \textcolor{tablegreen}{\CheckmarkBold} & \textcolor{tablegreen}{\CheckmarkBold} & \textcolor{tablegreen}{\CheckmarkBold} & \textcolor{tablegreen}{\CheckmarkBold} & \textcolor{tablegreen}{\CheckmarkBold} & \textcolor{tablegreen}{\CheckmarkBold} & \textbf{6.4K} & \textbf{6.0K} & 4.0K \\
\bottomrule
\end{tabular}}
\end{threeparttable}
\end{table*}

Hospital clinicians use electronic health records (EHRs) to review patient histories, place orders, and document care~\citep{wang2019ehrusage,wornow2023shaky}. Much of this work takes place through the EHR's graphical user interface (GUI).
To place a follow-up order, for example, a clinician may consult a prior report in one view and enter the new request in another. An agent assisting with this workflow must find the relevant information on screen and act on it in the patient's chart.

However, most medical agents still work through text or structured tool calls. Medical question answering tests reasoning from a clinical description~\citep{singhal2023large,lievin2024can}, while EHR benchmarks let agents query records or submit orders through structured requests~\citep{lee2022ehrsql,jiang2025medagentbench}.
This leaves open whether an agent can find the evidence on screen, choose the right control, and recognize when its action succeeds. Clinical GUI benchmarks and focused EHR training environments have begun to study these skills~\citep{anonymous2026physicianuibench,yu2026medcuabench,adtserapio2026ehrgym}. The challenge is to evaluate and train agents across the range of tasks clinicians perform through hospital EHRs.

In this work, we introduce \texttt{MedUIGym}, an interactive environment for evaluating and training GUI agents on clinical tasks in a simulated hospital EHR (Table~\ref{tab:related}). It brings together 15,739 tasks from 17 benchmark sources across six families: information retrieval, clinical decision making, cohort queries, clinical actions, clinical documentation, and outcome prediction. The tasks draw on patient-derived and synthetic records and range from reviewing one patient's treatment to identifying a cohort across many admissions.

In \texttt{MedUIGym}, agents read EHR screenshots, act with a mouse and keyboard, and observe the result on screen (Figure~\ref{fig:teaser}(b,c)). A native-tool interface also gives access to the same clinical records through database queries or FHIR requests.
Each source evaluator checks the requested result: an answer, an exported table, or a change to the chart such as a saved order or note. The environment collects trajectories in parallel and pairs each with its task score.

We benchmark twelve vision-language models on clinical GUI tasks. Several models perform well on tasks such as retrieving a laboratory result, but struggle when a workflow also requires choosing a treatment and documenting the decision. In the GUI trajectories we examine, common errors include selecting the wrong record or control. Greater reasoning effort helps on some queries but does not consistently improve performance on other clinical tasks.

We also use \texttt{MedUIGym} to train agents on these workflows. Collected GUI trajectories support offline learning, and task scores provide rewards for online reinforcement learning. Supervised fine-tuning improves in-domain performance for both Qwen3.5 models. Training with task rewards also improves their average scores on tasks from external clinical sources. Our contributions are as follows:

\begin{itemize}
\item \textbf{The first comprehensive environment for evaluating and training GUI agents on clinical tasks from multiple sources.}
Across six task families, agents review patient charts, identify cohorts across admissions, and enter orders and notes in a simulated hospital EHR.

\item \textbf{GUI interaction with verifiable task outcomes.}
We collect trajectories in parallel as agents read EHR screens and use the mouse and keyboard. These trajectories support offline learning. Scores for answers, exported tables, or chart changes serve as rewards for online training.

\item \textbf{Benchmarking and improving clinical GUI agents.}
Workflows that combine chart review, treatment decisions, and documentation remain difficult across models. Training Qwen3.5 agents in the EHR improves performance on both internal and external clinical tasks.
\end{itemize}

\section{Related Work}
\label{sec:related}

\begin{table*}[t]
\captionsetup{width=\linewidth,singlelinecheck=false}
\centering
\caption{\textbf{Dataset statistics}. Patient-derived and synthetic EHR data across 17 sources, with data types and sizes, patient counts, task categories, and training and test counts for the original collections and the \texttt{MedUIGym} leaderboard.}
\label{tab:data}
\begin{threeparttable}
\renewcommand\arraystretch{1.05}
\resizebox{\linewidth}{!}{%
\begin{tabular}{l|cccc|c|ccc|ccc}
\toprule
& \multicolumn{4}{c|}{\textbf{Data Sources}} & \multicolumn{4}{c|}{\textbf{Task Instances (all)}} & \multicolumn{3}{c}{\textbf{Tasks ({leader-board})}} \\
\cmidrule(lr){2-5} \cmidrule(lr){6-9} \cmidrule(lr){10-12}
\textbf{Dataset} & {Type} & {\#Patients} & {\#Table} & {\#Elements} & {Category} & {\#Train} & {\#Test} & {\#Total}  & {\#Train} & {\#Test} & {\#Total} \\
\midrule
\multicolumn{12}{l}{\textit{Training and Internal Validation (In-Distribution)}} \\
\midrule
MIMIC-III~\citep{johnson2016mimic} & Tabular & 1K & 17 & 1.4M & 7 & 9,318 & 1,122 & 10,440 & 931 & 421 & 1,352 \\
eICU / EHRSQL~\citep{pollard2018eicu,lee2022ehrsql} & Tabular & 1K & 10 & 1.5M & 7 & 9,270 & 1,117 & 10,387 & 872 & 411 & 1,283 \\
MIMICSQL / TREQS~\citep{wang2020text} & Tabular & 46.5K & 5 & 24.9M & 4 & 8,988 & 996 & 9,984 & 936 & 420 & 1,356 \\
EHR-Complex~\citep{qiao2026ehrcomplex} & Tabular & 364.6K & 31 & 886M & 12 & 48,092 & 3,915 & 52,007 & 673 & 346 & 1,019 \\
EHRStruct Synthea~\citep{yang2026ehrstruct} & Tabular & $\geq$1K & 7 & 59.6K & 11 & 550 & 550 & 1,100 & 550 & 500 & 1,050 \\
FHIR-AgentBench~\citep{lee2026fhiragentbench} & FHIR & 100 & -- & 928.9K & 11 & 2,098 & 833 & 2,931 & 1,000 & 398 & 1,398 \\
MedInsider~\citep{taha2026medinsider} & Tab\&Text & 420 & 1 & 10.2K & 5 & 420 & 420 & 840 & 420 & 420 & 840 \\
Synthetic Hospital~\citep{park2026synthetichospital} & Tab\&Text & 1.3K & 1 & 58.9K & 4 & 7,619 & 4,395 & 12,014 & 985 & 490 & 1,475 \\
\rowcolor{RoyalPurple!6} \textbf{\texttt{MedUIGym} (Internal)} & -- & $\geq$364.6K & 72 & 914.9M & 61 & 86,355 & 13,348 & 99,703 & 6,367 & 3,406 & 9,773 \\
\midrule
\multicolumn{12}{l}{\textit{External Validation (Out-of-Distribution)}} \\
\midrule
EHR-SeqSQL~\citep{ryu2024ehr} & Tabular & 1K & 17 & 1.4M & 6 & 6,375 & 2,820 & 9,195 & 951 & 455 & 1,406 \\
EHR-RobustGym~\citep{qiao2026ehrrobustgym} & Tabular & 364.6K & 31 & 886M & 6 & 8,956 & 2,016 & 10,972 & 1,000 & 446 & 1,446 \\
MedAgentBench v1~\citep{jiang2025medagentbench} & FHIR & 695$^1$ & -- & 786K & 10 & 0 & 300 & 300 & 0 & 300 & 300 \\
MedAgentBench v2~\citep{chen2026medagentbenchv2} & FHIR & 695$^1$ & -- & 786K & 10 & 0 & 300 & 300 & 0 & 300 & 300 \\
PhysicianBench~\citep{liu2026physicianbench} & FHIR & 108 & -- & 210.7K & 1 & 0 & 100 & 100 & 0 & 100 & 100 \\
MEMOIR~\citep{memoir2026} & Tab\&Text & 117 & 1 & 65.4K & 4 & 1,799 & 1,818 & 3,617 & 934 & 471 & 1,405 \\
AgentRx$^2$~\citep{aljorf2026agentrx} & Tab\&Text & 21.5K & 3 & 11M & 1 & 18,302 & 7,155 & 25,457 & 334 & 167 & 501 \\
MedAgentBoard (prediction)~\citep{zhu2025medagentboard} & Tab\&Text & 561 & 2 & 14.2K & 2 & 140 & 621 & 761 & 140 & 334 & 474 \\
MedAgentBoard (workflows)~\citep{zhu2025medagentboard} & Tabular & 461 & 2 & 2.2K & 3 & 0 & 100 & 100 & 0 & 34 & 34 \\
\rowcolor{RoyalPurple!6} \textbf{\texttt{MedUIGym} (External)} & -- & $\geq$364.6K & 56 & 899.5M & 43 & 35,572 & 15,230 & 50,802 & 3,359 & 2,607 & 5,966 \\
\midrule
\multicolumn{12}{l}{\textit{Overall}} \\
\midrule
\rowcolor{RoyalPurple!12} \textbf{\texttt{MedUIGym}} & -- & \textbf{$\geq$364.6K} & \textbf{80} & \textbf{927.0M} & \textbf{104} & \textbf{121,927} & \textbf{28,578} & \textbf{150,505} & \textbf{9,726} & \textbf{6,013} & \textbf{15,739} \\
\bottomrule
\end{tabular}%
}
\end{threeparttable}
\par\vspace{2pt}
\begin{minipage}{\textwidth}\footnotesize
\textsuperscript{1}\,MedAgentBench counts FHIR Patient resources.\quad \textsuperscript{2}\,We reconstruct AgentRx tasks using the authors' data splits and outcome definitions.
\end{minipage}
\end{table*}

\noindent\textbf{Clinical Reasoning in EHR Environments.}
Existing EHR benchmarks evaluate clinical agents on tasks ranging from information retrieval to clinical workflows.
EHRSQL and EHR-SeqSQL focus on retrieval and compositional database queries~\citep{lee2022ehrsql,ryu2024ehr}, while EHR-Complex extends these tasks to longitudinal, multi-table reasoning at patient and population levels~\citep{qiao2026ehrcomplex}.
MedAgentBench and its successor evaluate FHIR-based retrieval and clinical actions. MedAgentBench v2 guides subsequent interactions with memory derived from tasks each model previously failed~\citep{jiang2025medagentbench,chen2026medagentbenchv2}.
PhysicianBench combines clinical interpretation, treatment planning, and documentation~\citep{liu2026physicianbench}.
EHRSHOT and MIMIC-Extract focus on clinical outcome prediction from longitudinal patient records~\citep{wornow2023ehrshot,MIMICExtract}.
Complementing these benchmarks, \texttt{MedUIGym} emphasizes clinical computer use across diverse EHR tasks (Table~\ref{tab:related}).
Agents navigate EHR screens, retrieve and analyze patient information, make clinical decisions, and complete clinical actions and documentation using the mouse and keyboard.
We compare task completion through native tools and the GUI using the same clinical records and source-specific evaluators.

\noindent\textbf{Scalable and Interactive GUI Training Environments.}
Interactive environments support agent learning through repeated execution, feedback, and outcome verification.
MedAgentGym standardizes biomedical coding tasks for trajectory sampling and agent optimization~\citep{xu2025medagentgym}, while EHR-RobustGym supports training for accurate answering and evidence-grounded abstention under noisy clinical queries~\citep{qiao2026ehrrobustgym}.
PhysicianUIBench and MedCUA-Bench evaluate clinical GUI interaction~\citep{anonymous2026physicianuibench,yu2026medcuabench}. EHRGym supports training on EHR workflows~\citep{adtserapio2026ehrgym}, while HealthAdminBench evaluates agents on healthcare administration tasks~\citep{bedi2026healthadminbench}.
\texttt{MedUIGym} brings diverse clinical tasks, source-specific outcome verification, and parallel trajectory collection into one EHR environment.
It supports repeated sampling of successful and unsuccessful trajectories for agent training, with internal and external test sets for evaluating transfer across EHR data sources.

\section{MedUIGym}
\label{sec:method}

\begin{figure}[t]
\centering
\includegraphics[width=\linewidth]{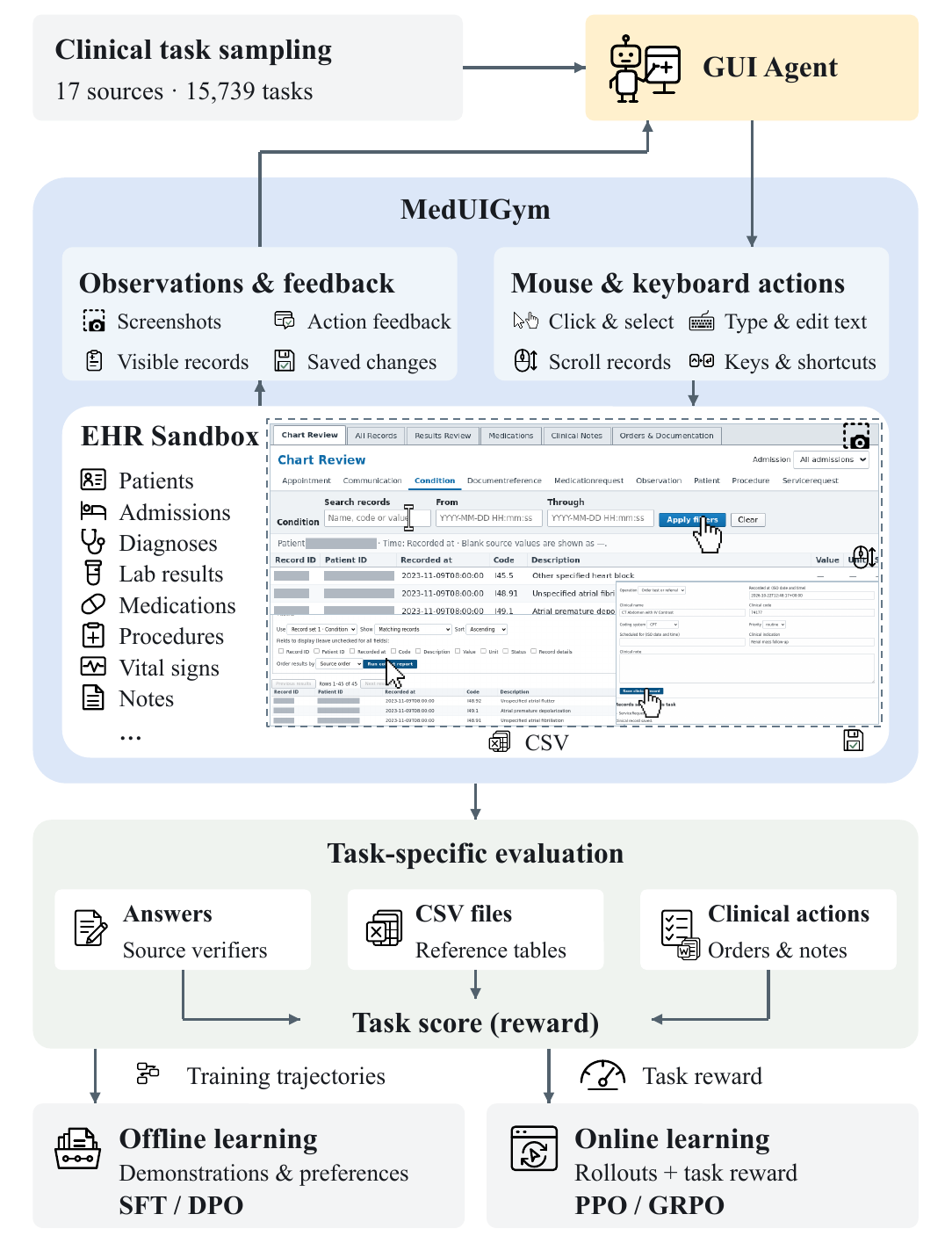}
\caption{\textbf{Clinical GUI interaction}. \texttt{MedUIGym} connects clinical instructions, screenshots, and GUI actions in an EHR environment. Agents click, type, scroll, and use keyboard shortcuts to complete clinical tasks with action feedback.}
\label{fig:overview}
\end{figure}

\subsection{Problem Formulation}
We formulate clinical GUI interaction as a structured problem-solving task: given a clinical instruction $x \in \mathcal{X}$, the agent generates a trajectory $c \in \mathcal{C}$ that produces a task output $y \in \mathcal{Y}$.
EHR screenshots and action feedback guide the agent's mouse and keyboard actions. For example, it can scroll through a patient's chart to review a prior CT, select a follow-up order, type its clinical indication, and check that the order appears in the chart after saving.
The output $y$ comprises the final answer and, when required, saved CSV files, orders, or clinical notes.
The task-specific evaluator $\mathcal{E}_x(c,y;y^*)$ scores the outcome against the reference answer or clinical requirements $y^*$, with partial credit where the source allows it.
Existing EHR benchmarks provide task descriptions, references, and evaluation procedures, but generally do not supply GUI trajectories. For a given task, \texttt{MedUIGym} can sample trajectories $c^{(i)}$ in parallel, pairing each with its output $y^{(i)}$ and source score. Successful and unsuccessful attempts provide examples for agent training.

\subsection{Data Construction: From Individual Datasets to Unified Benchmark}
\label{sec:method-data}

\textbf{Task and Data Identification.}
\texttt{MedUIGym} focuses on verifiable clinical tasks that require interaction with electronic health records.
\emph{Clinically}, we prioritize tasks grounded in patient records and established EHR benchmarks, including EHRSQL~\citep{lee2022ehrsql}, EHR-Complex~\citep{qiao2026ehrcomplex}, and MedAgentBench~\citep{jiang2025medagentbench}.
\emph{Computationally}, we organize these tasks into six families: information retrieval, clinical decision making, cohort queries, clinical actions, clinical documentation, and outcome prediction.
These tasks range from retrieving a laboratory result and identifying a patient cohort to reviewing longitudinal records and entering a clinical order.
For outcome prediction, the agent uses the available clinical records to predict the outcome for an individual patient.

\textbf{Task Instructions and Reference Outcomes.}
To standardize tasks across sources, each instance in \texttt{MedUIGym} includes (1) a problem description, (2) a reference answer or clinical requirements, and (3) the corresponding EHR data and evaluation procedure.
For query tasks, the evaluator compares the submitted answer with the result of a database query or the answer supplied by the source.
For clinical actions, the evaluator checks whether the requested orders, notes, or other changes appear in the patient record.
We correct task instructions and evaluators where needed, retaining the source requirements and scoring criteria for most tasks, including numerical tolerances, answer equivalence, and clinical checkpoints.
For the 34 MedAgentBoard EHR extraction tasks~\citep{zhu2025medagentboard}, originally assessed by hand, the agent saves the requested records as a CSV file. Our evaluator compares its contents with the complete reference table.

\textbf{Data Statistics.}
Table~\ref{tab:data} summarizes 150,505 tasks from 17 benchmark sources. The leaderboard includes 15,739 tasks across 102 of the 104 source-specific categories.
We retain up to 1,000 training and 500 test tasks per source. Eight sources supply training and test tasks and nine supply external tasks. EHR-Complex (internal) and EHR-RobustGym (external) use the same MIMIC-IV v3.1 database from PhysioNet to answer different clinical questions~\citep{johnson2024mimiciv,johnson2023mimiciv,pollard2026physionet}.

\subsection{GUI Environment: From Native Tools to GUI Interaction}
\label{sec:method-env}

\begin{table*}[t]
\captionsetup{width=\linewidth,singlelinecheck=false}
\centering

\caption{\textbf{Pass@1 and ATC results}. LLM performance across reasoning settings on eight in-distribution sources of \texttt{MedUIGym}.}

\small
\renewcommand\arraystretch{0.85}
\setlength{\tabcolsep}{3.620pt}
\setlength{\aboverulesep}{0.2ex}
\setlength{\belowrulesep}{0.3ex}
\begin{tabular}{lc|*{8}{c}|c}
\toprule
\textbf{Model} & \textbf{Thinking} & \textbf{MIMIC.} & \textbf{eICU} & \textbf{TREQS} & \textbf{EHR-C.} & \textbf{ES-S} & \textbf{FHIR} & \textbf{MedIns.} & \textbf{SynHos.} & \textbf{Avg.} \\
 & \textit{(effort)} & SR & SR & SR & SR & SR & SR & ATC & SR & Score \\
\midrule
\multicolumn{11}{l}{\textit{API-based Proprietary LLMs}} \\
\midrule
\modelicon{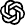}\texttt{GPT-6.1 Terra}$^\dagger$ & max & 76.2 & 85.4 & 73.8 & 70.6 & 88.0 & 90.0 & 100.0 & 28.6 & 76.6 \\
\modelicon{openai}\texttt{GPT-5.4 Mini} & low & 34.9 & 38.4 & 5.5 & 11.3 & 42.8 & 48.0 & 8.0 & 7.3 & 24.5 \\
\modelicon{openai}\texttt{GPT-5.4 Mini} & high & 37.3 & 36.7 & 10.5 & 15.3 & 45.6 & 54.3 & 0.0 & 7.8 & 25.9 \\
\modelicon{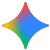}\texttt{Gemini 3.6 Flash}$^\dagger$ & high & 73.8 & 78.0 & 57.1 & 26.5 & 88.0 & 72.5 & 99.1 & 20.4 & 64.4 \\
\modelicon{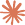}\texttt{Claude Sonnet 5.5}$^\dagger$ & max & 76.2 & 63.4 & 64.3 & 44.1 & 68.0 & 75.0 & 92.9 & 8.2 & 61.5 \\
\midrule
\multicolumn{11}{l}{\textit{OSS ($>$ 100B parameters)}} \\
\midrule
\modelicon{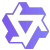}\texttt{Qwen3.5-397B-A17B} & $\times$ & 26.8 & 26.0 & 7.9 & 3.8 & 34.4 & 35.4 & 23.3 & 5.1 & 20.3 \\
\modelicon{qwen}\texttt{Qwen3.5-397B-A17B} & $\checkmark$ & 33.7 & 32.4 & 9.3 & 1.7 & 41.8 & 43.0 & 21.1 & 4.3 & 23.4 \\
\modelicon{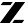}\texttt{GLM-5.3-Flash} & high & 49.2 & 55.2 & 46.4 & 7.2 & 66.8 & 64.8 & 69.3 & 18.0 & 47.1 \\
\modelicon{zai}\texttt{GLM-5.3-Flash} & max & 56.8 & \textbf{64.7} & \textbf{60.2} & 11.3 & 71.4 & 62.1 & 79.6 & 18.2 & 53.0 \\
\modelicon{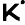}\texttt{Kimi K3} & low & 38.5 & 43.1 & 32.4 & 11.8 & 73.4 & 59.5 & 88.3 & 13.5 & 45.1 \\
\modelicon{kimi}\texttt{Kimi K3} & high & 49.6 & 55.7 & 43.1 & 25.1 & \textbf{79.4} & 64.8 & \textbf{93.2} & 16.9 & 53.5 \\
\modelicon{kimi}\texttt{Kimi K3} & max & 50.1 & 52.1 & 37.1 & 24.6 & 75.8 & 62.3 & 92.1 & 14.9 & 51.1 \\
\modelicon{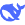}\texttt{DeepSeek-V4.1-Flash-1017} & high & \textbf{58.0} & 61.3 & 45.2 & \textbf{32.1} & 69.0 & \textbf{66.8} & 84.7 & \textbf{18.4} & \textbf{54.4} \\
\modelicon{deepseek}\texttt{DeepSeek-V4.1-Flash-1017} & max & 56.8 & 57.4 & 44.3 & 29.5 & 63.6 & 62.1 & 31.9 & 15.3 & 45.1 \\
\midrule
\multicolumn{11}{l}{\textit{OSS ($\leq$ 100B parameters)}} \\
\midrule
\modelicon{qwen}\texttt{Qwen3.6-27B} & $\times$ & 4.5 & 3.6 & 0.0 & 0.9 & 8.8 & 6.3 & 0.1 & 1.8 & 3.3 \\
\modelicon{qwen}\texttt{Qwen3.6-27B} & $\checkmark$ & 7.6 & 9.0 & 0.7 & 0.6 & 13.4 & 7.5 & 0.1 & 3.3 & 5.3 \\
\modelicon{qwen}\texttt{Qwen3.8-27B} & $\times$ & 11.2 & 12.2 & 1.0 & 1.4 & 14.6 & 10.3 & 11.6 & 5.7 & 8.5 \\
\modelicon{qwen}\texttt{Qwen3.8-27B} & $\checkmark$ & 16.2 & 16.8 & 2.6 & 3.8 & 26.2 & 14.6 & 10.4 & 8.4 & 12.4 \\
\bottomrule
\end{tabular}%
\par\vspace{2pt}
\begin{minipage}{\linewidth}\footnotesize
\textbf{Bold} indicates the best result in each column among models evaluated on the full test set. $^\dagger$ Models use the same stratified 10\% subset (340 tasks).
\end{minipage}%
\label{tab:main_results}
\end{table*}

\textbf{Isolated and Interactive EHR Environment.}
The real desktop applications in OSWorld~\citep{xie2024osworld} and the clinical workflows in PhysicianUIBench~\citep{anonymous2026physicianuibench} inspired our EHR interface (Figure~\ref{fig:overview}).
AI researchers with medical backgrounds helped design and refine the interface to reflect clinical practice, including patient search, chart review, cohort queries, order entry, and documentation.
Patient records, encounters, laboratory results, and medications remain linked to their source data, providing the clinical context for task completion.
Each task begins in an independent EHR environment. Orders and notes created during one interaction are confined to that task, while the underlying clinical data remain unchanged.
The GUI agent works from screenshots and action feedback, using the mouse and keyboard. Native tools expose the same clinical records through database queries or FHIR requests.
Both interfaces use the same clinical objectives and evaluators (Section~\ref{subsec:native-gui}). In Figure~\ref{fig:teaser}(b), the reference query selects patients with right upper extremity cellulitis at their first admission and links their next admission to a mean corpuscular hemoglobin (MCH) test. Each patient counts once, even if several tests are recorded during that admission.

\textbf{Interactive Feedback.}
\texttt{MedUIGym} connects agents to the interface through the standard OpenAI tools format.
The system prompt specifies task rules, tool schemas define action names and parameters, and the task instruction specifies the required final answer.
Actions issued in one model response are executed in sequence. The agent receives feedback for each executed action and a single screenshot after the last one.
It can then revisit a record, adjust a query, or correct an incomplete order. Interaction ends when the agent submits, gives up, or reaches the turn or action limit.
On submission, the source evaluator assesses the final answer and any required files or changes to the patient record.

\textbf{Efficient Trajectory Collection.}
\texttt{MedUIGym} supports parallel interaction across independent tasks and repeated sampling for a given task.
Each trajectory links the screenshots and actions used to obtain clinical evidence with the resulting feedback and task outcome.
Only trajectories from the training split are used for agent training. On test tasks, the trajectories identify the patient records the agent consulted and the controls it used to obtain the final answer.

\section{Evaluations with MedUIGym}
\label{sec:exp}

\subsection{Experimental Setup}\label{subsec:setup}
\textbf{Agent Scaffolds.}
We use a common scaffold to evaluate clinical GUI agents.
Interactions within \texttt{MedUIGym} are modeled as a Partially Observable Markov Decision Process (POMDP)~\citep{kaelbling1998planning}, focusing on clinical tasks $p \in \mathcal{P}$.
The agent selects GUI actions $a_{t+1} \in \mathcal{A}$ under policy $\pi_t$ using the latest model-visible screenshot $o_t \in \mathcal{O}$ and retained interaction history.
A single response may contain several mouse or keyboard actions. The agent has up to 100 model turns and 100 actions per task, including submission.

\textbf{Tasks and Datasets.}
The internal split comprises 6,367 training tasks and 3,406 test tasks across eight datasets:
(1) MIMIC-III~\citep{johnson2016mimic},
(2) eICU~\citep{pollard2018eicu} from EHRSQL~\citep{lee2022ehrsql},
(3) MIMICSQL/TREQS~\citep{wang2020text},
(4) EHR-Complex~\citep{qiao2026ehrcomplex},
(5) EHRStruct Synthea,
(6) FHIR-AgentBench,
(7) MedInsider, and
(8) Synthetic Hospital.
We conduct \emph{out-of-distribution} evaluation on 2,607 test tasks from EHR-SeqSQL, EHR-RobustGym, MedAgentBench v1 and v2, PhysicianBench, MEMOIR, AgentRx, and the prediction and workflow tasks of MedAgentBoard.
We train on tasks drawn from the eight internal benchmark sources (Table~\ref{tab:data}).

\begin{table*}[t]
\captionsetup{width=\linewidth,singlelinecheck=false}
\centering

\caption{\textbf{Fine-tuning results}. Pass@1 and ATC scores on eight in-distribution sources of \texttt{MedUIGym}, with gains over the base models.}

\small
\renewcommand\arraystretch{0.85}
\setlength{\tabcolsep}{6.275pt}
\setlength{\aboverulesep}{0.2ex}
\setlength{\belowrulesep}{0.3ex}
\begin{tabular}{l|*{8}{c}|cc}
\toprule
\textbf{Datasets ($\rightarrow$)} & \textbf{MIMIC.} & \textbf{eICU} & \textbf{TREQS} & \textbf{EHR-C.} & \textbf{ES-S} & \textbf{FHIR} & \textbf{MedIns.} & \textbf{SynHos.} & \textbf{Avg.} & \textcolor{OliveGreen}{\textbf{$\Delta_{\mathrm{Base}}$}} \\
\textbf{Base ($\downarrow$) / Metrics ($\rightarrow$)} & SR & SR & SR & SR & SR & SR & ATC & SR & Score & (pp) \\
\midrule
\rowcolor{Gray!16} \modelicon{qwen}\texttt{Qwen3.5-4B} & 9.7 & 12.9 & 0.0 & 0.6 & 20.2 & 12.6 & 0.2 & 4.5 & 7.6 & -- \\
\quad +SFT & 18.5 & 15.1 & 2.1 & 0.6 & 22.2 & 7.8 & 14.9 & 1.8 & 10.4 & \textcolor{OliveGreen}{(\emph{+2.80})} \\
\quad \quad +DPO & 26.1 & 24.3 & 3.8 & 3.5 & 31.8 & 11.1 & 11.3 & 0.6 & 14.1 & \textcolor{OliveGreen}{(\emph{+6.48})} \\
\quad \quad +PPO & 17.1 & 17.0 & 1.9 & 1.7 & 22.0 & 6.8 & 21.3 & 1.6 & 11.2 & \textcolor{OliveGreen}{(\emph{+3.60})} \\
\rowcolor{Peach!16} \quad \quad +GRPO & 19.7 & 18.5 & 2.1 & 2.0 & 31.2 & 8.5 & 41.2 & 4.3 & 16.0 & \textcolor{OliveGreen}{(\emph{+8.37})} \\
\midrule
\rowcolor{Gray!16} \modelicon{qwen}\texttt{Qwen3.5-9B} & 8.8 & 10.0 & 0.5 & 1.2 & 17.4 & 12.6 & 0.1 & 3.7 & 6.8 & -- \\
\quad +SFT & 21.1 & 21.2 & 3.8 & 4.0 & 35.4 & 12.3 & 9.2 & 2.7 & 13.7 & \textcolor{OliveGreen}{(\emph{+6.95})} \\
\quad \quad +DPO & 25.7 & 24.3 & 6.4 & 2.0 & 52.0 & 16.8 & 14.2 & 2.9 & 18.0 & \textcolor{OliveGreen}{(\emph{+11.27})} \\
\quad \quad +PPO & 22.6 & 20.4 & 4.5 & 2.9 & 39.4 & 14.1 & 13.6 & 4.3 & 15.2 & \textcolor{OliveGreen}{(\emph{+8.46})} \\
\rowcolor{SeaGreen!16} \quad \quad +GRPO & 24.2 & 24.6 & 4.3 & 4.0 & 41.6 & 14.3 & 39.6 & 7.1 & 20.0 & \textcolor{OliveGreen}{(\emph{+13.21})} \\
\bottomrule
\end{tabular}%
\label{tab:training_results}
\end{table*}

\textbf{Baselines.}
We evaluate twelve models, including ten baselines and two fine-tuning backbones.
Seven baselines share the GUI scaffold and limits on all 6,013 test tasks.
We vary reasoning effort for \texttt{GPT-5.4 Mini} (low, high), \texttt{Kimi K3} (low, high, max), \texttt{GLM-5.3-Flash} (high, max), and \texttt{DeepSeek-V4.1-Flash-1017} (high, max).
We test \texttt{Qwen3.5-397B-A17B}, \texttt{Qwen3.6-27B}, and \texttt{Qwen3.8-27B} with thinking enabled and disabled.
We evaluate the three remaining baselines ($\dagger$) on a fixed stratified subset of 340 ID and 261 OOD tasks.
Given the high API costs of these commercial frontier models, we sample about 10\% per source across all test categories.

\textbf{Evaluation Metrics.}
We evaluate one attempt per task and adopt \emph{success rate (SR)} as the primary metric, using the source-specific evaluators for final answers and completed clinical actions.
AgentRx and MedAgentBoard prediction use \emph{accuracy (Acc)}.
MedInsider uses \emph{Aligned Task Completion (ATC)}~\citep{taha2026medinsider}, the fraction of required targets completed.
Tasks ending in a tool-use error or another unscored outcome receive zero.
Avg.\ Score gives each source equal weight. We average task scores within each source, scale them by 100, and report the mean of the unrounded source scores separately for the eight ID and nine OOD sources.

\subsection{Evaluation Results}

\textbf{Mortality Prediction Distinguishes the ID Leaders on External Tasks.}
Among the fifteen configurations evaluated on both full test sets, \texttt{DeepSeek-V4.1-Flash-1017} and \texttt{Kimi K3} rank first and second on ID tasks at high reasoning effort (54.4 and 53.5). Their order reverses on OOD tasks (41.4 and 47.8, Tables~\ref{tab:main_results} and~\ref{tab:ood_results}).
The largest difference is on AgentRx, where the models predict in-hospital mortality from the first 48 hours of ICU records. Accuracy on AgentRx is 31.1\% for DeepSeek and 68.3\% for Kimi, accounting for 4.1 of the 6.4 points between their average OOD scores.

\textbf{Completing All Parts of a Clinical Workflow Remains Difficult.}
On the full test sets, the best SR reaches 87.0\% on MedAgentBench v1 and 66.8\% on FHIR-AgentBench, compared with 15.5\% on EHR-RobustGym and 3.0\% on PhysicianBench.
The same contrast appears for GPT-6.1 Terra on the shared subset: SR is 100\% on MedAgentBench v1 and 10\% on PhysicianBench. MedAgentBench v1 includes tasks such as retrieving the latest magnesium level. PhysicianBench requires all parts of a workflow to succeed. In a case of drug-resistant urinary tract infection, the agent must assess renal function, select an antibiotic, and document its safety before the patient receives chemotherapy.

\textbf{Higher Reasoning Effort Can Improve Queries While Lowering Prediction Accuracy.}
Raising \texttt{GLM-5.3-Flash} from high to max effort increases SR on TREQS from 46.4\% to 60.2\%, while accuracy on AgentRx falls from 61.7\% to 31.7\%. Its ID average rises from 47.1 to 53.0 and its OOD average falls from 44.3 to 42.7.
All three Qwen baselines improve with thinking enabled. For \texttt{Qwen3.5-397B-A17B}, the ID and OOD averages rise from 20.3 to 23.4 and from 20.1 to 27.5, respectively. For \texttt{Kimi K3}, high effort gives the highest average score on both the internal and external sources.

\begin{figure*}[t]
\centering
\begin{minipage}[t]{0.60\linewidth}
\centering
\includegraphics[width=\linewidth]{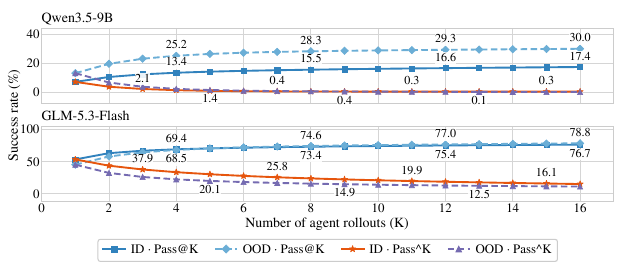}
\par\smallskip
{\small (a) \textbf{Inference-Time Scaling}}
\end{minipage}\hfill
\begin{minipage}[t]{0.39\linewidth}
\centering
\includegraphics[width=\linewidth]{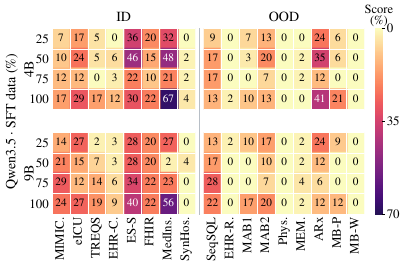}
\par\smallskip
{\small (b) \textbf{Training-Data Scaling}}
\end{minipage}
\caption{\textbf{Inference-time and training-data scaling}. (a)~Pass@K and \PassK{} for Qwen3.5-9B and GLM-5.3-Flash on ID and OOD tasks. (b)~ID and OOD source scores for Qwen3.5-4B and Qwen3.5-9B using different fractions of nearly 4,000 clinical GUI trajectories.}
\label{fig:scaling}

\medskip
\begin{minipage}[b]{0.342\linewidth}
\centering
\includegraphics[width=\linewidth]{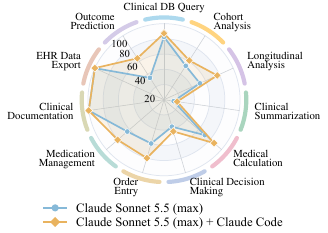}
\caption{\textbf{Claude Code on clinical tasks}.}
\label{fig:harness}
\end{minipage}\hfill
\begin{minipage}[b]{0.319\linewidth}
\centering
\includegraphics[width=\linewidth]{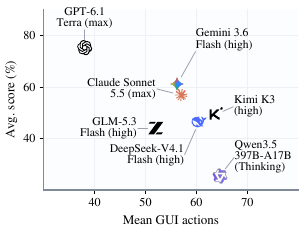}
\caption{\textbf{Task completion efficiency}.}
\label{fig:efficiency}
\end{minipage}\hfill
\begin{minipage}[b]{0.329\linewidth}
\centering
\includegraphics[width=\linewidth]{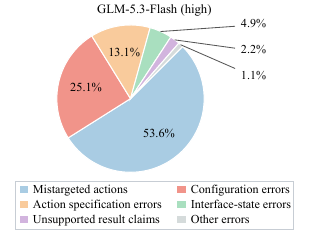}
\caption{\textbf{GUI interaction errors}.}
\label{fig:error}
\end{minipage}
\end{figure*}

\textbf{Scaffold Choice Can Improve Success Rates and Reduce GUI Actions.}
Claude Sonnet 5.5 (max) operates the same EHR controls through Claude Code's tool-use loop~\citep{anthropic2025claudecode}. SR rises from 55.3\% to 64.6\% on 559 tasks with binary evaluation, with the largest gains in outcome prediction and longitudinal analysis (Figure~\ref{fig:harness}).
On the 289 tasks both scaffolds solve, the median falls from 23 GUI actions to 17. In one EHR-SeqSQL case, both correctly compared mean arterial pressure across an ICU stay. With Claude Code, one search of the ICU flowsheet retrieved all four readings. Sonnet then verified the stay dates and reported both changes because two readings shared the final timestamp. It completed the task in 11 actions, compared with 20 under the common scaffold.

\begin{table*}[t]
\captionsetup{width=\linewidth,singlelinecheck=false}
\centering
\caption{\textbf{Generalization to external sources}. Pass@1 scores for zero-shot and fine-tuned models on nine sources of \texttt{MedUIGym}.}

\small
\renewcommand\arraystretch{0.85}
\setlength{\tabcolsep}{3.022pt}
\setlength{\aboverulesep}{0.2ex}
\setlength{\belowrulesep}{0.3ex}
\begin{tabular}{lc|*{9}{c}|c}
\toprule
\textbf{Model} & \textbf{Thinking} & \textbf{SeqSQL} & \textbf{EHR-R.} & \textbf{MAB1} & \textbf{MAB2} & \textbf{Phys.} & \textbf{MEM.} & \textbf{ARx} & \textbf{MB-P} & \textbf{MB-W} & \textbf{Avg.} \\
 & \textit{(effort)} & SR & SR & SR & SR & SR & SR & Acc & Acc & SR & Score \\
\midrule
\multicolumn{12}{l}{\textit{API-based Proprietary LLMs}} \\
\midrule
\modelicon{openai}\texttt{GPT-6.1 Terra}$^\dagger$ & max & 73.9 & 68.9 & 100.0 & 86.7 & 10.0 & 48.9 & 88.2 & 90.9 & 100.0 & 74.2 \\
\modelicon{openai}\texttt{GPT-5.4 Mini} & low & 26.6 & 3.6 & 55.3 & 43.3 & 0.0 & 13.8 & \textbf{89.2} & 62.3 & 0.0 & 32.7 \\
\modelicon{openai}\texttt{GPT-5.4 Mini} & high & 36.3 & 6.5 & 46.7 & 21.3 & 0.0 & 18.5 & 81.4 & 6.0 & 0.0 & 24.1 \\
\modelicon{gemini}\texttt{Gemini 3.6 Flash}$^\dagger$ & high & 67.4 & 26.7 & 96.7 & 30.0 & 0.0 & 51.1 & 88.2 & 63.6 & 100.0 & 58.2 \\
\modelicon{claude}\texttt{Claude Sonnet 5.5}$^\dagger$ & max & 73.9 & 42.2 & 90.0 & 26.7 & 0.0 & 40.4 & 17.6 & 84.8 & 100.0 & 52.9 \\
\midrule
\multicolumn{12}{l}{\textit{OSS ($>$ 100B parameters)}} \\
\midrule
\modelicon{qwen}\texttt{Qwen3.5-397B-A17B} & $\times$ & 17.8 & 3.1 & 43.0 & 19.3 & 0.0 & 12.5 & 44.3 & 40.7 & 0.0 & 20.1 \\
\modelicon{qwen}\texttt{Qwen3.5-397B-A17B} & $\checkmark$ & 27.9 & 3.4 & 57.0 & 21.0 & 0.0 & 12.1 & 76.6 & 49.4 & 0.0 & 27.5 \\
\modelicon{kimi}\texttt{Kimi K3} & low & 38.9 & 11.7 & 75.3 & 53.3 & 1.0 & 37.4 & 73.7 & 67.4 & 29.4 & 43.1 \\
\modelicon{kimi}\texttt{Kimi K3} & high & 50.3 & \textbf{15.5} & 83.7 & 47.7 & \textbf{3.0} & \textbf{41.6} & 68.3 & 72.8 & \textbf{47.1} & \textbf{47.8} \\
\modelicon{kimi}\texttt{Kimi K3} & max & 51.2 & 14.1 & 80.0 & 40.0 & 1.0 & 39.9 & 64.7 & 76.3 & 44.1 & 45.7 \\
\modelicon{zai}\texttt{GLM-5.3-Flash} & high & 52.1 & 9.0 & 81.7 & \textbf{60.0} & \textbf{3.0} & 35.7 & 61.7 & 68.9 & 26.5 & 44.3 \\
\modelicon{zai}\texttt{GLM-5.3-Flash} & max & \textbf{58.5} & 10.8 & 85.0 & 50.0 & 1.0 & 38.6 & 31.7 & \textbf{76.6} & 32.4 & 42.7 \\
\modelicon{deepseek}\texttt{DeepSeek-V4.1-Flash-1017} & high & 49.2 & 11.0 & 82.3 & 52.7 & \textbf{3.0} & 35.0 & 31.1 & 60.8 & \textbf{47.1} & 41.4 \\
\modelicon{deepseek}\texttt{DeepSeek-V4.1-Flash-1017} & max & 52.5 & 10.1 & \textbf{87.0} & 35.3 & 1.0 & 36.3 & 7.8 & 73.7 & 41.2 & 38.3 \\
\midrule
\multicolumn{12}{l}{\textit{OSS ($\leq$ 100B parameters)}} \\
\midrule
\modelicon{qwen}\texttt{Qwen3.6-27B} & $\times$ & 0.9 & 0.0 & 4.7 & 10.7 & 0.0 & 1.1 & 19.8 & 16.2 & 0.0 & 5.9 \\
\modelicon{qwen}\texttt{Qwen3.6-27B} & $\checkmark$ & 0.2 & 0.2 & 8.7 & 10.7 & 0.0 & 0.8 & 31.7 & 23.4 & 0.0 & 8.4 \\
\modelicon{qwen}\texttt{Qwen3.8-27B} & $\times$ & 5.3 & 0.0 & 15.7 & 22.0 & 0.0 & 1.7 & 69.5 & 44.9 & 0.0 & 17.7 \\
\modelicon{qwen}\texttt{Qwen3.8-27B} & $\checkmark$ & 9.9 & 0.2 & 28.0 & 17.0 & 0.0 & 4.5 & 73.1 & 63.8 & 0.0 & 21.8 \\
\midrule
\rowcolor{Gray!16} \modelicon{qwen}\texttt{Qwen3.5-4B} & $\checkmark$ & 0.4 & 0.0 & 16.3 & 33.3 & 0.0 & 0.4 & 44.3 & 23.1 & 0.0 & 13.1 \\
\hspace{1.2em}\quad \textbf{+SFT} & $\checkmark$ & 9.7 & 0.7 & 4.3 & 12.7 & 0.0 & 0.4 & 68.3 & 13.5 & 0.0 & 12.2 \\
\hspace{1.2em}\quad \quad \textbf{+DPO} & $\checkmark$ & 18.0 & 1.3 & 4.0 & 5.3 & 0.0 & 1.5 & 59.3 & 16.8 & 0.0 & 11.8 \\
\hspace{1.2em}\quad \quad \textbf{+PPO} & $\checkmark$ & 9.7 & 0.7 & 3.7 & 14.0 & 0.0 & 0.6 & 70.1 & 15.9 & 0.0 & 12.7 \\
\rowcolor{Peach!16} \hspace{1.2em}\quad \quad \textbf{+GRPO} & $\checkmark$ & 12.3 & 0.9 & 7.7 & 33.0 & 0.0 & 2.8 & 76.6 & 33.8 & 0.0 & 18.6 \\
\midrule
\rowcolor{Gray!16} \modelicon{qwen}\texttt{Qwen3.5-9B} & $\checkmark$ & 0.0 & 0.0 & 9.0 & 24.0 & 0.0 & 0.6 & 51.5 & 32.6 & 0.0 & 13.1 \\
\hspace{1.2em}\quad \textbf{+SFT} & $\checkmark$ & 15.8 & 0.4 & 11.0 & 13.7 & 0.0 & 0.8 & 85.0 & 41.3 & 0.0 & 18.7 \\
\hspace{1.2em}\quad \quad \textbf{+DPO} & $\checkmark$ & 19.6 & 0.2 & 13.0 & 12.3 & 0.0 & 2.3 & 84.4 & 35.3 & 0.0 & 18.6 \\
\hspace{1.2em}\quad \quad \textbf{+PPO} & $\checkmark$ & 16.7 & 0.7 & 10.0 & 13.0 & 0.0 & 0.8 & 82.0 & 46.7 & 0.0 & 18.9 \\
\rowcolor{SeaGreen!16} \hspace{1.2em}\quad \quad \textbf{+GRPO} & $\checkmark$ & 16.3 & 0.4 & 13.7 & 25.3 & 0.0 & 1.5 & 84.4 & 57.8 & 0.0 & 22.2 \\
\bottomrule
\end{tabular}%
\par\vspace{2pt}
\begin{minipage}{\linewidth}\footnotesize
\textbf{Bold} indicates the best result in each column among models evaluated on the full test set. $^\dagger$ Models use the same stratified 10\% subset (261 tasks).
\end{minipage}%
\label{tab:ood_results}
\end{table*}

\subsection{Native-Tool and GUI Task Completion}
\label{subsec:native-gui}
On the 601 tasks evaluated on both interfaces using the same source criteria, \texttt{GLM-5.3-Flash} (high) solves 173 only with native tools and 81 only through the GUI (ATC $=1$ for MedInsider).
Moving from native tools to the GUI lowers the source-averaged score from 64.7 to 44.7 on ID tasks and from 69.3 to 43.5 on OOD tasks.
Native tools score higher on fourteen sources. EHR-Complex SR falls from 64.7\% to 5.9\%, while MedAgentBench v1 SR rises from 53.3\% to 80.0\%.
For the 396 tasks with identical instructions, the average score across the 13 sources is 63.8 for native tools and 42.8 for the GUI. An agent can specify search conditions directly in a native query. In \texttt{MedUIGym}, it must also check which chart filters remain active when it starts a new search.
On a MEMOIR task, a native query returned contraceptive implant records dating back to 2012. The GUI agent also sought the full contraceptive history, but its search for mestranol retained an earlier 2018--2020 filter. It later cleared the dates but did not repeat the search. The two answers earned the same score although the records retrieved from the chart differed.

\section{Training and Analysis with MedUIGym}
\label{sec:medcopilot}

\subsection{Agent Fine-tuning with Trajectory Sampling}
\label{sec:sft}

\textbf{Training Setup.}
Successful GUI trajectories from a pool of 2,000 internal training tasks provide supervision for \texttt{Qwen3.5-4B} and \texttt{Qwen3.5-9B}~\citep{qwen2026qwen35}.
A separate set of 2,000 tasks supplies one preference pair per task for direct preference optimization (DPO)~\citep{rafailov2023direct}.
The two groups cover the same 59 source-specific categories. On the second group, proximal policy optimization (PPO)~\citep{schulman2017proximal} and group relative policy optimization (GRPO)~\citep{shao2024deepseekmath} learn from rewards assigned by the source evaluators.
For each backbone, DPO, PPO, and GRPO start from the same SFT checkpoint.

\textbf{Supervised Fine-Tuning Improves Internal Scores at Both Model Sizes.}
SFT raises the internal average score of Qwen3.5-4B from 7.6 to 10.4 and that of Qwen3.5-9B from 6.8 to 13.7 (Table~\ref{tab:training_results}).
For Qwen3.5-9B, success rate increases from 17.4\% to 35.4\% on EHRStruct Synthea and from 10.0\% to 21.2\% on eICU.
On FHIR-AgentBench and Synthetic Hospital, SR falls for both models after SFT.

\textbf{One Clinical Source Accounts for the ID Lead of GRPO over DPO.}
DPO raises the ID average from the SFT scores of 10.4 and 13.7 to 14.1 and 18.0 for the 4B and 9B models, respectively.
PPO reaches ID averages of 11.2 and 15.2, while GRPO reaches 16.0 and 20.0.
MedInsider accounts for the difference. With the 9B model, GRPO completes 39.6\% of the requested targets on average, compared with 14.2\% under DPO. DPO achieves higher mean SR across the other seven sources for both backbones. On EHRStruct Synthea, for example, the 9B model reaches 52.0\% SR with DPO compared with 41.6\% with GRPO.

\subsection{Inference-Time and Training-Data Scaling}\label{subsec:scale}

\textbf{More Attempts Solve More Tasks, but Few Are Solved Every Time.}
We vary the number of candidate trajectories $K$ for \texttt{Qwen3.5-9B} and \texttt{GLM-5.3-Flash} (Figure~\ref{fig:scaling}(a)).
\emph{Pass@K}~\citep{chen2021evaluating} measures the fraction of tasks solved by at least one of the $K$ candidates, while \emph{\PassK{}}~\citep{yao2024tau} requires all $K$ to succeed.
We sample 16 independent candidates per task and average source scores separately for 340 ID and 261 OOD tasks. For MedInsider, success requires completing all requested targets (ATC $=1$).
For GLM-5.3-Flash on OOD tasks, Pass@K rises from 45.4\% at $K=1$ to 78.8\% at $K=16$, while \PassK{} falls to 11.3\%.

\textbf{The Best SFT Checkpoint Varies by Clinical Source.}
For Figure~\ref{fig:scaling}(b), we train each Qwen backbone once on four successive blocks of 1,000 tasks, evaluating each checkpoint on the same test subset.
On EHRStruct Synthea, the 9B model improves from 28.0\% to 40.0\% SR, while the 4B model peaks at the second checkpoint. On AgentRx, 9B accuracy falls from 23.5\% to 5.9\%, then rises to 11.8\%.

\subsection{Generalization, Efficiency, and Error Analysis}
\label{sec:exp-agent}

\textbf{Transfer after Supervised Fine-Tuning Depends on Model Size.}
On external sources, SFT raises the average score for Qwen3.5-9B from 13.1 to 18.7, while the 4B average falls from 13.1 to 12.2 (Table~\ref{tab:ood_results}).
Averaged across external sources, the share of tasks with higher scores after SFT is 10.8\% for 9B and 6.9\% for 4B. Scores decline on 5.2\% and 7.8\%, respectively. GRPO achieves the highest external averages for both backbones (18.6 and 22.2 for 4B and 9B). On MedAgentBoard prediction, 9B accuracy rises from 41.3\% after SFT to 57.8\% after GRPO.
PPO yields smaller gains over SFT on external tasks for both backbones, while DPO improves mainly on the internal sources.

\textbf{Similar GUI Action Counts Can Yield Different Task Scores.}
Figure~\ref{fig:efficiency} compares seven models on the 601-task subset. We give each of the 17 sources equal weight in the score and report the mean number of GUI actions per task.
A click and a text entry count as separate actions even when the model issues both in one response. GPT-6.1 Terra achieves the highest score (75.3) with the fewest actions per task (38.1 on average).
Kimi K3 and Qwen3.5-397B-A17B average a similar number of actions (63.7 and 64.7), but Kimi's score is 23.6 points higher on the same clinical tasks.

\textbf{Agents Misread Values and Mistarget Actions.}
A MedAgentBoard chart recorded 30 breaths per minute. GLM-5.3-Flash reported the upper reference limit of 18 as the patient's respiratory rate.
In other tasks, the agent searched ICD procedure records for a CPT billing event and recorded a daily maximum of 50\,mg for a regimen allowing at most 45\,mg.
GPT-6.1 Terra reviewed the screenshots, actions, and EHR feedback for the 601 tasks attempted by GLM-5.3-Flash (high). GUI errors appeared on 156 tasks. Figure~\ref{fig:error} counts each error type once per task, giving 183 observations. The most frequent errors were actions directed at the wrong control or record (98 tasks) and incorrectly configured queries or forms (46 tasks).
The agent earned full credit on 53 of the 156 tasks. In one case, it clicked the clinical notes field while trying to save a vaccine booster order. The form stayed open until the agent clicked Save. The EHR then confirmed the order.

\section{Conclusion}
\label{sec:conclusion}

Clinical GUI agents must find relevant information on screen and act on it in the patient's chart.
We introduced \texttt{MedUIGym} to evaluate and train agents on this work in a simulated hospital EHR, with 15,739 tasks from 17 sources across six task families.
Agents read screenshots and use a mouse and keyboard. The paired native tools use the same records and scoring criteria for answers and completed clinical actions.
Our evaluation of twelve vision-language models shows that workflows combining chart review, treatment decisions, and documentation remain difficult, and gains from greater reasoning effort vary across tasks.
GUI trajectories support offline learning, and task scores provide rewards for online reinforcement learning.
After training, both Qwen3.5 models achieve higher average scores on clinical GUI tasks from internal and external sources.

{
    \small
    \bibliographystyle{ieeenat_fullname}
    \bibliography{main}
}

\end{document}